\documentclass[acmsmall]{acmart}

\AtBeginDocument{%
  \providecommand\BibTeX{{%
    \normalfont B\kern-0.5em{\scshape i\kern-0.25em b}\kern-0.8em\TeX}}}

\renewcommand\footnotetextcopyrightpermission[1]{}

\setcopyright{none}

\usepackage{siunitx}
\usepackage[utf8]{inputenc}
\usepackage{mathptmx} % This is Times font
\usepackage{fancyhdr}
\usepackage[normalem]{ulem}
\usepackage{microtype}
\usepackage{enumitem}

\usepackage{graphicx}
\graphicspath{{plots/}}
\usepackage{algorithm,algorithmic}

\usepackage{subfig}
\usepackage{color}
\usepackage{wrapfig}
\usepackage{soul}
\usepackage{threeparttable}

\newcommand{\ignore}[1]{}

\setlist{leftmargin=1em}
\newcommand{\Sec}[1]{Section~\ref{#1}}
\newcommand{\Fig}[1]{Figure~\ref{#1}}

\renewcommand{\paragraph}[1]{\noindent\textbf{#1}}

\title{Exploiting Parallelism Opportunities with Deep Learning Frameworks}

\author{Yu Emma Wang}
\email{ywang03@g.harvard.edu}
\affiliation{
    \institution{Harvard University}
    \streetaddress{33 Oxford St}
    \city{Cambridge}
    \state{MA}
    }

\author{Carole-Jean Wu}
\email{carolejeanwu@fb.com}
\affiliation{\institution{Facebook}}

\author{Xiaodong Wang}
\email{xdwang@fb.com}
\affiliation{\institution{Facebook}}

\author{Kim Hazelwood}

\affiliation{\institution{Facebook}}

\author{David Brooks}
\email{dbrooks@eecs.harvard.edu}
\affiliation{\institution{Harvard University}}

\begin{document}

\begin{abstract}
State-of-the-art machine learning frameworks support a wide variety of design features to enable a flexible machine learning programming interface and to ease the programmability burden on machine learning developers. Identifying and using a performance-optimal setting in feature-rich frameworks, however, involves a non-trivial amount of performance profiling efforts and often relies on domain-specific knowledge. This paper takes a deep dive into analyzing the performance impact of key design features in a machine learning framework and quantifies the role of parallelism. The observations and insights distill into a simple set of guidelines that one can use to achieve much higher training and inference speedup. Across a diverse set of real-world deep learning models, the evaluation results show that the proposed performance tuning guidelines outperform the Intel and TensorFlow recommended settings by 1.29$\times$ and 1.34$\times$, respectively.
\end{abstract}

\maketitle
\thispagestyle{plain}
\pagestyle{plain}

\section{Introduction}

The wide adoption of deep learning (DL) tasks has spawned a plethora of domain-specific frameworks to help improve development efficiency and to enable a flexible path from research to production deployment for machine learning (ML). Notable ML frameworks include Caffe/Caffe2~\cite{jia2014caffe}, PyTorch~\cite{ketkar2017introduction}, TensorFlow~\cite{abadi2016tensorflow}, and MXNet~\cite{chen2015mxnet}.
These frameworks provide high-level APIs as the building blocks of DL models.
The ML frameworks can significantly reduce the prototyping cycle because of the substantial (re)use of optimized libraries and greatly improve the productivity of developers building end-to-end DL models.
In addition to programmability benefits, the frameworks also provide many DL-specific optimizations to improve performance and performance portability across software stacks and new hardware systems.
The net result is an explosion in the development of ever more complex DL models and a concomitant increase in the performance cost of inference.

Performance is especially important for deep learning inference.
When trained DL models are deployed into at-scale production serving, performance optimization for inference can affect datacenter resource efficiency since inference is sensitive to both throughput and latency~\cite{reddi2019mlperf,hazelwood2018applied,sriraman2019softsku}.
Inference performance also affects whether a model can be deployed to performance- and energy-constrained mobile and edge systems~\cite{wu2019machine}.
An important research question this paper aims to address is 
---{\it what is the performance cost of high-level abstraction or programmability in modern feature-rich DL frameworks?}.
A further question examined in the paper is \textit{whether and by how much can we reduce this ``programmability tax'' by tuning the complex set of design choices available in current ML frameworks}.
Answering both questions requires a comprehensive understanding of framework features.

Previous work have examined the performance of the modern DL frameworks and focused on cross-framework comparison~\cite{bahrampour2016comparative,jain2019performance,shi2016benchmarking}. These studies help machine learning model developers and engineers to make informed decisions about framework choices. 
However, there lacks in-depth, focused performance characterization studies that quantify performance impacts of key framework design features, explaining the underlying performance root cause of each design feature within a framework.
These design features include a rich set of parallelism opportunities within and between DL model layers and across input data, for example, across the batch dimension. Furthermore, this paper also examines key choices available to users include back-end math kernels, threading libraries, and scheduling policies (Sections~\ref{sec:scheduler} to~\ref{sec:library}).

\begin{figure}[t]
    \centering
    \subfloat{\includegraphics[width=0.7\columnwidth]{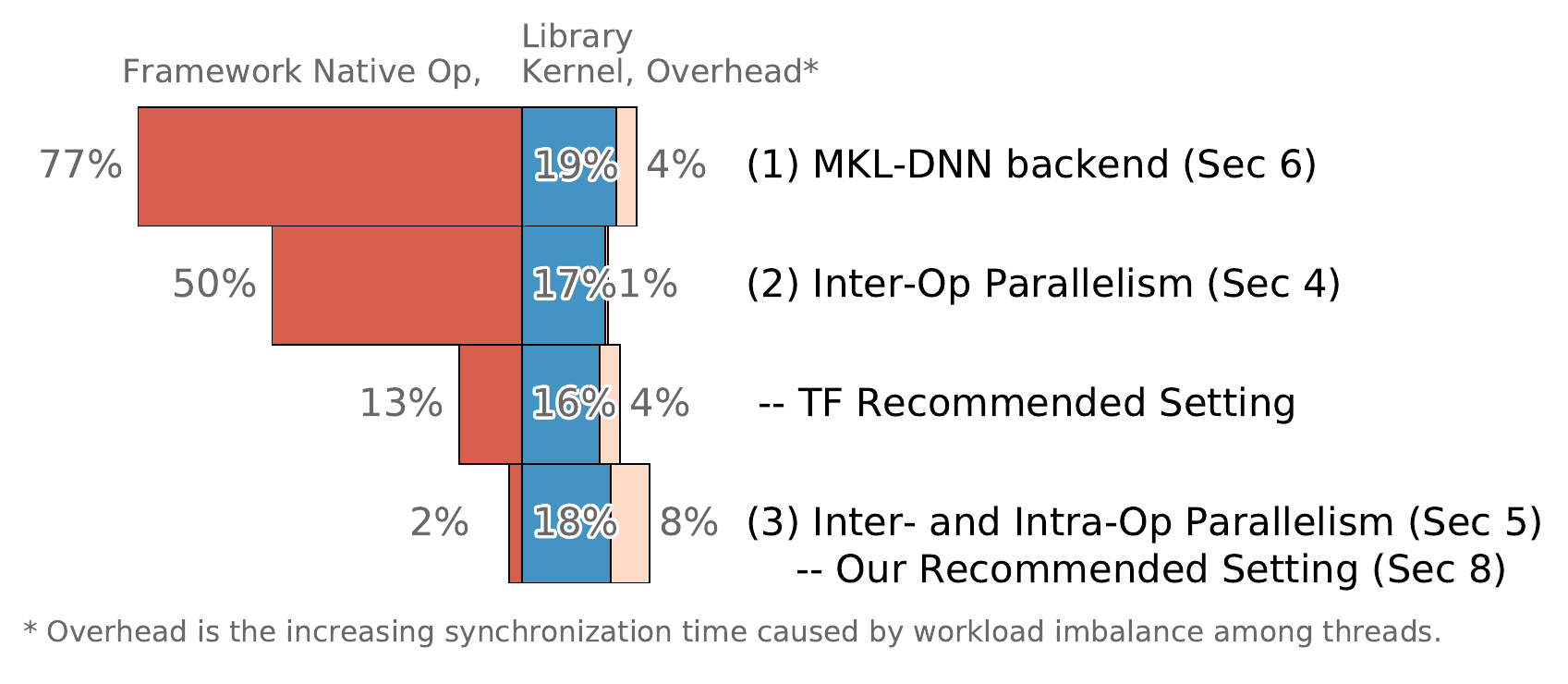}}

    \caption{Time breakdown for Inception v3.
    }
    
    \label{fig:motivation}
\end{figure}

\Fig{fig:motivation} uses Inception as an example to illustrate the performance impact of important framework features across the programming abstraction\footnote{Performance breakdown for Inception v3 running on a dual-socket CPU platform is presented. Methodological detail is in Section~\ref{sec:methodology}.}. Overall, the performance of Inception can be improved by \textbf{3.6$\times$} with properly tuned, optimized framework configurations.
Determining performance-optimal framework knobs, however, requires expert knowledge.
In this example, tuning the inter-operator parallelism speeds up the performance of native operators by 1.54$\times$, leading to a speedup of \textbf{1.47$\times$} for the whole model.
Exploiting intra-operator parallelism further speeds up inference performance by another 25$\times$, translating to \textbf{2.4$\times$} performance improvement for the entire model.
Using the performance tuning guideline described in this work can speed up
the native operators by 6.5$\times$, leading to an overall model performance improvement of \textbf{3.6$\times$}.
Compared to the recommended TensorFlow setting~\cite{tensorflowperformance}, our tuning guideline speeds up the whole model by 1.15$\times$.
The evaluation results across a wide collection of DL workloads demonstrate consistent performance speedup. For some DL workloads, over 2$\times$ performance improvement is observed.
Selecting the optimal performance setting is not straightforward, even for CPU platforms that serve a large, diverse collection of DL use cases in production datacenter fleets. The significant gap for performance improvement motivates the in-depth studies presented in this paper.

This paper provides three major contributions: 
\begin{itemize}
    \item We provide a detailed analysis to reason about and quantify performance implications of fundamental framework design features on CPU platforms,
    including scheduling mechanisms, operator designs, library back ends, and parallelism mechanisms.
We find that the programmability tax ranges from 63\% to 1.3\%.
    \item Built upon the insights from this performance analysis, we propose simple guidelines for tuning framework parameters to maximize parallelism and reduce framework overheads. 
    \item We demonstrate the usability of our approach by integrating the proposed methodology into TensorFlow. The code is made available on GitHub\footnote{\url{https://github.com/Emma926/mcbench}}. The proposed settings match the globally optimal performance and
    outperform the suggested settings from Intel~\cite{inteltips} and TensorFlow~\cite{tensorflowperformance} performance guides by 1.29$\times$ and 1.34$\times$, respectively, across a set of real-world DL models, including several from the MLPerf suite~\cite{mlperf}.
\end{itemize}

\section{Framework Design Overview}
\label{sec:overview}

\begin{figure*}[t]
\begin{center}
\includegraphics[width=1\columnwidth]{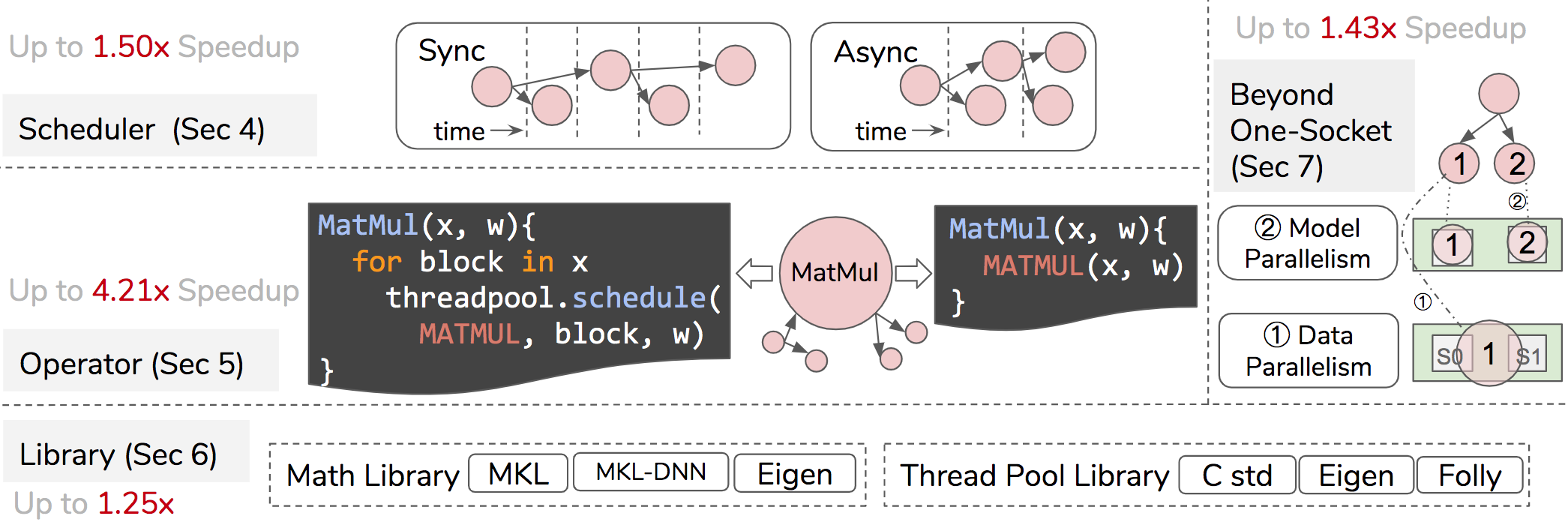}
\caption{An overview of the framework design features studied in this work.}
\label{fig:overview}
\end{center}
\end{figure*}

Deep learning frameworks
have lowered the programming effort for DL researchers and decreased prototyping time for new models.
Abstraction provided by these frameworks hides some design choices, such as run-time scheduling and actual implementation of operators, which may not be noticeable by users 
but are important for performance.
In this section, we describe how deep learning framework design choices (\Sec{sec:overview:features}) exploit parallelism opportunities exposed in deep learning workloads (\Sec{sec:overview:opportunities}),
and overview our framework parameter tuning methodology (\Sec{sec:overview:tuning}).
We also discuss related work.
While this work focuses on inference, our analysis and takeaways can be applied for training as well. Thus we demonstrate both inference and training.

Our performance analysis focuses on CPUs, which are the most widely-used platforms serving ML workloads in datacenters~\cite{hazelwood2018applied,gupta2019architectural}. Performance improvement directly translates into capacity efficiency improvement.
As we will see, DL frameworks offer a large set of tradeoffs to exploit parallelism that lead to significant optimization opportunities on CPUs.
On the other hand, frameworks for accelerators such as GPUs expose less performance knobs for users, because mapping compute kernels to accelerators introduces less variability than mapping to general purpose processors.
As developers focusing on accelerating a handful of compute kernels for deep learning, the kernel computation will be increasingly faster and the framework overhead will take larger fraction.
At this point, efficient designs of framework is especially important, and the lessons learnt from parallelizing CPU frameworks can be transferred to frameworks for accelerators.

\subsection{Design Features}
\label{sec:overview:features}

\Fig{fig:overview} presents the stack of DL frameworks and design features that we study.
This work focuses on frameworks with opaque operators, i.e., operators or kernels that are manually written and tuned by engineers. It is a design adopted by popular DL frameworks such as TensorFlow, Keras, Pytorch and Caffe2.
Frameworks such as Tensor Comprehensions~\cite{vasilache2018tensor},
TVM~\cite{chen2018tvm},
and
Julia~\cite{bezanson2017julia} do not belong to this category since they generate kernels. The configurations they explore are usually not exposed to model developers at the framework level such as the ones in \Fig{fig:overview}.
Thus, the designs of those frameworks differ in important ways that put them out of the scope of this paper.

The design feature study in this work distinguishes itself from previous work that focus on framework-level comparison without revealing the root causes of the performance difference, including \cite{jain2019performance,shi2016benchmarking,bahrampour2016comparative}.
This work emphasizes that the performance difference of frameworks originates from their different design choices, such as the ones in \Fig{fig:overview}.
This applies to different versions of the same framework, such as TensorFlow 1 and 2.
The insights from analyzing TensorFlow 1 are applicable to TensorFlow 2 because they share the same design features and the performance difference is caused by their different choices.

\paragraph{Scheduler}
Here ``scheduler'' refers to the operator scheduler, not the process scheduler of the OS.
It takes a computational graph representing the DL workload and schedules its operators based on dependencies and hardware resources.
Two common approaches are synchronous and asynchronous scheduling.
Synchronous scheduling schedules one operator at a time.
Asynchronous scheduling schedules all operators in ready state,
such that the operators can execute in parallel if hardware resources are available.
In the example of \Fig{fig:overview}, asynchronous scheduling is faster than synchronous scheduling, assuming unlimited hardware units.
However as we will show in \Sec{sec:scheduler}, given limited hardware resources, the optimal mechanism usually falls between the two extremes.

\paragraph{Operator}
Frameworks include both native operators and operators based on library kernels.
The way operators make use of kernels can have a surprisingly large impact on performance.
For example,
\Fig{fig:overview} shows two potential implementations of a MatMul operator based on library kernel MATMUL, to implement the MatMul operator at the framework level.
The right one passes arguments as is to MATMUL.
The left one splits matrix $x$ into smaller blocks and passes each block to a thread in a thread pool.
We will show in \Sec{sec:op:intraop} that the latter performs better because it parallelizes data preparation before entering the MATMUL kernel.

\paragraph{Library}
Mathematical libraries provide efficient parallel implementations of common kernels.
We study three widely-used libraries: MKL, MKL-DNN and Eigen.
A thread pool manages a number of threads that execute tasks upon request.
Besides the thread pools used by math libraries,
DL frameworks use additional thread pools to parallelize computations outside of the math kernels.
We study thread pool implementations in the C standard library, Eigen, and Folly.

\paragraph{Beyond One-Socket}
Parallelism
mechanisms need to be applied based on workloads.
Common mechanisms include
data and model parallelism.

\subsection{Parallelism Opportunities}
\label{sec:overview:opportunities}

A DL workload can be expressed with a computational graph, where a node represents an operator, and an edge indicates the dataflow dependencies between operators~\cite{abadi2016tensorflow}.
DL workloads expose the parallelism within an operator (intra-operator), between operators (inter-operator),
and among requests.
Efficient framework designs should exploit such opportunities.

\subsubsection{Parallelism within an Operator}

Operators manipulate tensors, i.e., $n$-dimensional arrays.
The parallelism within an operator (intra-operator parallelism) can be exploited with 
the following techniques.

\paragraph{SIMD}
The use of single instruction multiple data (SIMD) architectures, e.g., Intel's AVX instructions~\cite{lomont2011introduction}, is implemented in mathematics libraries.
The MKL, MKL-DNN, and Eigen libraries can use AVX2 and AVX512 instructions.

\paragraph{Multi-Threading}
Multi-threading is implemented at the operator and library levels.
For example, MKL uses OpenMP for multi-threading.
At the operator level, a
framework may have a separate thread pool for further parallelism.

\paragraph{Data Parallelism}
Data parallelism splits one batch of data
into multiple smaller batches.
Thus it can improve performance of large-batch workloads.

\subsubsection{Parallelism between Operators}
\label{sec:design:interop}
The parallelism across operators (inter-operator parallelism) can be exploited by scheduling and model parallelism.

\paragraph{Scheduling}
Asynchronous scheduling is
to place independent operators on different hardware units such as CPU threads, GPU stream multi-processors, or processing elements in accelerators.
This way independent operators can execute in parallel.

\paragraph{Model Parallelism}
Model parallelism is realized by scheduling different operators (or the same operator after splitting along the model size dimension) on different hardware sockets or nodes,
such as distributing large embedding tables across hardware nodes~\cite{eisenman2018bandana}.

\paragraph{Model Pipelining}
Model pipelining is a special case of model parallelism~\cite{gpipe}.
One hardware node receives data from a previous node and operates on it while the previous node is computing the next training step.
In contrast to data parallelism, model parallelism lowers the memory requirement for large models.
This work does not study model pipelining.

\subsubsection{Parallelism among Requests}

Parallelism among requests can be exploited by batching, to transform request-level parallelism to intra-op parallelism.
For example, multiple image classification requests can be combined and executed in a single session, such that the number of requests is mapped to the batch size dimension.

\subsection{Framework Parameter Tuning}
\label{sec:overview:tuning}

Based on our analysis of design features and parallelism opportunities,
we reduce the number of design features needing to be selected from five (scheduling mechanism, operator design, math library, thread pool library, parallelism mechanism) to one, the number of asynchronous scheduling thread pools.
Other features, such as operator parallelism, follow from that choice.
We propose simple guidelines for tuning framework features based on a model's inter-operator parallelism, as reflected in its computational graph.
To demonstrate their usability, we integrate the guidelines with TensorFlow, and achieve 1.29$\times$ and 1.34$\times$ speedup over Intel~\cite{inteltips} and TensorFlow~\cite{tensorflowperformance} recommended settings, respectively.
We also achieve the same average performance as with globally optimal settings and 95\% of globally optimal performance in the worse case.
We have developed a TensorFlow plugin that sets framework parameters automatically, and it is available on GitHub.\footnote{\url{https://github.com/Emma926/mcbench}}
Details are in \Sec{sec:tuning}.

Previous work proposed to tune TensorFlow parameters automatically~\cite{hasabnis2018auto}, which treats the tuning process as a black box and therefore does not explain how parameters affect performance.
Our work differs in three ways.
First, our tuning method is supported by strong analysis.
A deep understanding of framework designs and the root causes of performance difference makes our tuning method intuitive and lightweight.
Second, performance with our worst-case settings differs by less than 5\% from the \textit{global optimum} obtained by exhaustive search,
while previous work~\cite{hasabnis2018auto} reports large performance degradation.
Finally, 
the robustness of our guidelines,
is validated on a different set of real-world DL workloads using a state-of-the-art two-socket system.
Our evaluation highlights the importance of framework parameter tuning
for state-of-the-art translation and recommendation models.
\section{Experimental Setup}
\label{sec:methodology}
Our scripts and workloads are available on GitHub, together with the TensorFlow plugin.

\paragraph{CPU Platforms}
We use three Intel Skylake CPU platforms, \textit{small}, \textit{large}, and \textit{large.2}.
\textit{large.2} contains two sockets of \textit{large} with a peak bi-directional bandwidth of 120 GB/s.
\textit{large} and \textit{large.2} represent widely-used datacenter servers.
\textit{small} has fewer cores; we use it to eliminate threading overhead for certain studies. 
We use \textit{large} and \textit{small} for most of the analysis where the performance tuning guidelines are summarized, and \textit{large.2} for the evaluation of the guidelines. 
\textit{small} has 32 fused multiply-add (FMA) units per core, while \textit{large} has 64 per core.
Although all of the platforms support hyperthreading, each core has only one set of FMA units,
which limits the benefits of
hyperthreading if both hyperthreads need FMA units.
The specifications are summarized in Table~\ref{table:cpu}.
\textit{large} and \textit{large.2} are Amazon Web Services \texttt{m5.metal} instances.
Unless otherwise specified, we use the \textit{large} platform.

\begin{threeparttable}[t]

\small
\begin{center}
\begin{tabular}{|c|c|c|c|c|c|c|c|c|c|}
\multicolumn{1}{c}{} & \multicolumn{1}{c}{SKU} &\multicolumn{1}{c}{Cores} & \multicolumn{1}{c}{TFLOPS} & \multicolumn{1}{c}{ Freq}& \multicolumn{1}{c}{LLC}\\ 
\hline
\textit{small} & i7-6700k & 4 & 0.423* & \SI{4}{GHz} & \SI{8}{MB} \\
\hline
\textit{large} & Platinum 8175M & 24 & 1.64*  & \SI{2.5}{GHz} & \SI{33}{MB} \\
\hline 
\end{tabular}
\begin{tablenotes}
\item[*]Estimated with GeekBench v4~\cite{geekbench}.
\end{tablenotes}
\end{center}
\caption{Hardware platforms under study. An additional platform is $large.2$ that contains two sockets of $large$ with a 120GB/s bi-directional UPI bandwidth at peak.}

\label{table:cpu}
\end{threeparttable}

\paragraph{Frameworks}
Because TensorFlow supports all features, we use TensorFlow v1.13 with the MKL-DNN back end, unless otherwise specified.
Conducting the same experiments with PyTorch (Caffe2 module) shows similar trends, so in this paper we focus on TensorFlow.
%MKL and MKL-DNN use OpenMP to parallelize the kernels.
We set thread affinity to prioritize binding one software thread with one physical core~\cite{inteltips}.

\paragraph{Workloads}
We use a set of production-size deep learning models, including three from MLPerf~\cite{mlperf} (ResNet-50~\cite{he2016deep}, Transformer~\cite{vaswani2017attention}, neural collaborative filtering (NCF)~\cite{he2017neural}), as well as DenseNet~\cite{huang2017densely}, SqueezeNet~\cite{iandola2016squeezenet}, Inception~\cite{szegedy2016rethinking}, GoogLeNet~\cite{szegedy2015going}, CaffeNet~\cite{jia2014caffe}, ResNext~\cite{xie2017aggregated}, and Google's Wide \& Deep Learning model~\cite{widedeep}.
To deeply understand the design features, we use micro-benchmarks such as matrix multiplication.
We use a subset of the aforementioned models to focus our evaluation on the respective design features (Sections~\ref{sec:scheduler} and~\ref{sec:operator}).
We hold out all the non-vision models for \Sec{sec:tuning} to evaluate the proposed method.

\paragraph{Methodology}
Our profiling methodology enables thorough analysis.
We produce time breakdowns (stack bars) for individual CPU cores using Linux's \texttt{perf} \texttt{record} and profiling one core at a time.
Using floating-point performance counters, we measure performance as floating-point operations per second (FLOPS).
We trace execution with performance counters by sampling instructions per cycle (IPC) every few milliseconds and ordering the samples by time stamps.
The \texttt{perf} \texttt{stat} command with the \texttt{topdown} option produces the top-down breakdown.
LLC misses, memory, and UPI traffic come from the corresponding performance counters.

\paragraph{Terminology}
Here are some terms used often later on.

\begin{itemize}
    \item \textbf{MKL Threads}: the threads for MKL and MKL-DNN.
    Abbreviated intra-op threads. 
    \item \textbf{Inter-Operator Pools}: the independent thread pools in a framework, the size of each set by intra-op threads. Abbreviated inter-op pools, or pools. 
\end{itemize}

\section{Scheduling Mechanism}
\label{sec:scheduler}

\begin{figure}[t]
    \centering

    \subfloat[Synchronous\label{fig:sched:example:a} ]{\includegraphics[width=0.18\columnwidth]{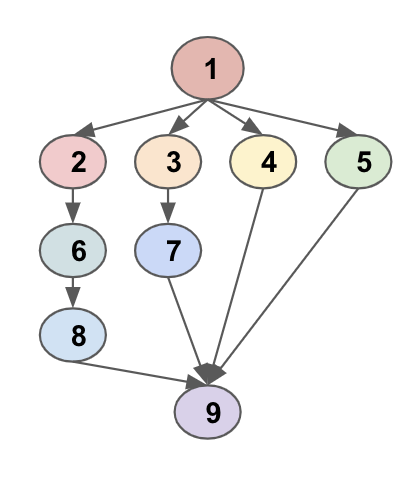}}
    \subfloat[Asynchronous\label{fig:sched:example:b} ]{\includegraphics[width=0.2\columnwidth]{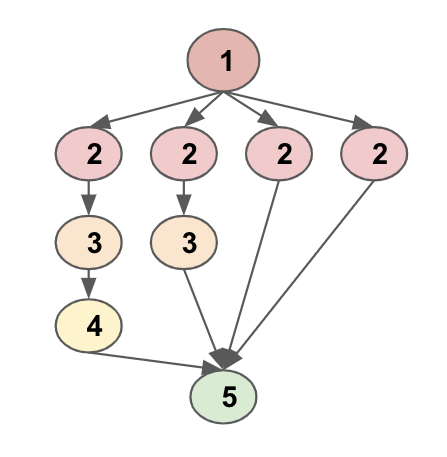}}
    \subfloat[Thread Pools\label{fig:sched:example:c}]{\includegraphics[width=0.2\columnwidth]{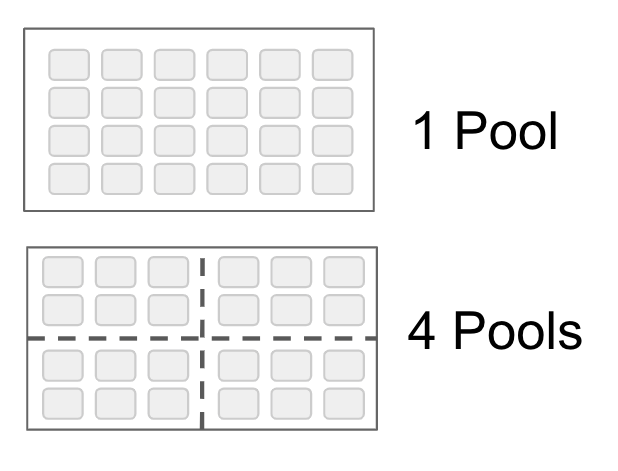}}

    \caption{Examples of (a) synchronous scheduling, (b) asynchronous scheduling, and (c) using one and four thread pools, with the same total hardware resources.}

    \label{fig:sched:example}
\end{figure}

A deep learning model is expressed using a computational graph that represents the data flow between operators.
At run time, operator scheduling exposes optimization opportunities,
such as scheduling independent operators simultaneously.
In this section we study the trade-offs of using such inter-operator parallelism\footnote{The experiments are conducted with the real-world workloads implemented in Caffe2, because the Inception architecture is important for this study, and the Caffe2 model zoo makes it more convenient to use Inception.}.
We show that not all models benefit from asynchronous scheduling, and that the best setting depends on a model's inter-operator parallelism, quantified by the width of its computational graph.

\Fig{fig:sched:example} shows examples of synchronous and asynchronous scheduling of an Inception module~\cite{szegedy2016rethinking}, and the use of one and four thread pools.
Scheduling of an operator is to submit the job to a thread pool.
Sometimes asynchronous scheduling, i.e., running multiple operators simultaneously, can improve performance.
For the simple example shown, scheduling one operator at a time takes nine steps to finish (\Fig{fig:sched:example:a}); scheduling four operators at a time reduces the steps to five (\Fig{fig:sched:example:b}).
One simple implementation is to create several thread pools of the same size to share the computing hardware (\Fig{fig:sched:example:c}),
and to schedule independent operators asynchronously to the thread pools.
This design is adopted by popular DL frameworks.
In TensorFlow, the number of asynchronous thread pools is called the number of inter-operator threads. Caffe2 calls it the asynchronous thread pool size.
In this paper, we refer to it as the inter-op pools, as opposed to intra-op threads.
We will show that synchronous scheduling is beneficial in both single-socket and multi-socket systems.
The best performance is achieved by balancing intra- and inter-operator parallelism.

\begin{figure}[t]
    \centering
    \subfloat{\includegraphics[width=0.7\columnwidth]{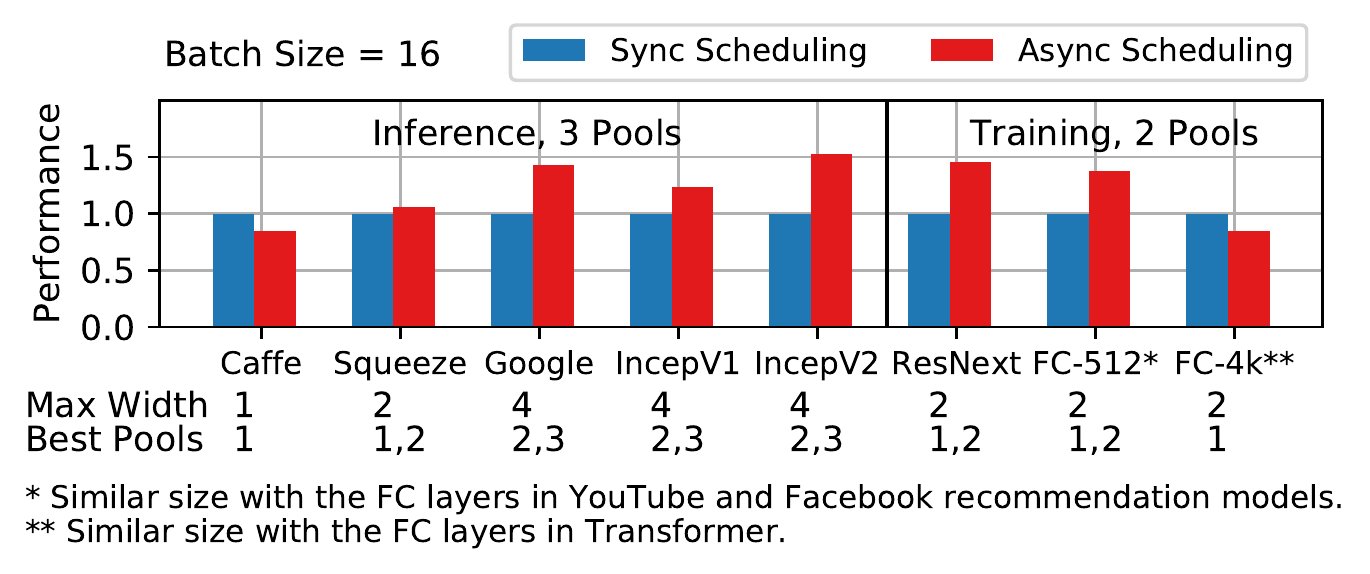}}

    \caption{(Bar Chart) The speedup of using asynchronous scheduling over synchronous.
    (Table) The maximum computational graph width and best numbers of thread pools.
    Workloads with more branches benefit from more pools.
    }

    \label{fig:sched:speedup}
\end{figure}

\subsection{Datacenter Platform Performance}
\label{sec:sched:performance}

We show that the best number of thread pools is no more than the maximum number of parallel operators for a model.
We use the \textit{large} platform in Table~\ref{table:cpu}.

\Fig{fig:sched:speedup}'s bar chart shows the speedup of asynchronous scheduling on different production-size inference and training workloads.
The baseline is synchronous scheduling, using one thread pool of size 24.
Inference uses three thread pools, each with 8 threads; training uses two pools, each with 12 threads.
Workloads benefit from asynchronous scheduling differently.
Inception v1 and v2, GoogLeNet, ResNet, and FC-512 speed up more than others.

The performance difference is because of the intrinsic inter-op parallelism of the models, which is quantified by the width, or the number of branches, of their computational graphs.
It measures the number of operators that can be scheduled in parallel.
The table at the bottom of \Fig{fig:sched:speedup} summarizes the maximum graph width and the best numbers of pools for varying batch sizes.
We distinguish between inference and training workloads because the computational graphs of training workloads contain gradient and sum weight operators, which doubles the number of parallel operators. 
An intrinsic model limitation is that the best numbers of pools (for varying batch sizes) do not exceed the maximum graph width.
The best number of pools varies based on batch sizes.
Large batches increase the best number of pools for inference, but decrease it for training.
That is because the parallel operators for training, gradient and sum weight, become imbalanced with large batches.
Gradient becomes much compute-intensive than sum weight.
Allocating computing resources evenly for the two hurts performance.

\subsection{Inception v2 Case Study}
\label{sec:scheduler:inception}
To highlight parallelism opportunities at the intra- and inter-op levels, we use Inception v2 as an example.
Its model architecture contains operator  branches that can execute in parallel.
In the baseline implementations that either schedule each branch naively to one CPU core or schedule one operator to all CPU cores, workload imbalance and synchronization overhead significantly reduce performance.
We show that synchronization overhead can be mitigated by choosing the number\slash size of thread pools to better balance intra- and inter-operator parallelism.
We use the \textit{small} platform in Table~\ref{table:cpu}, as it enables an exhaustive study of possible cases.

\begin{figure}[t]
    \centering
    \subfloat{\includegraphics[width=0.8\columnwidth]{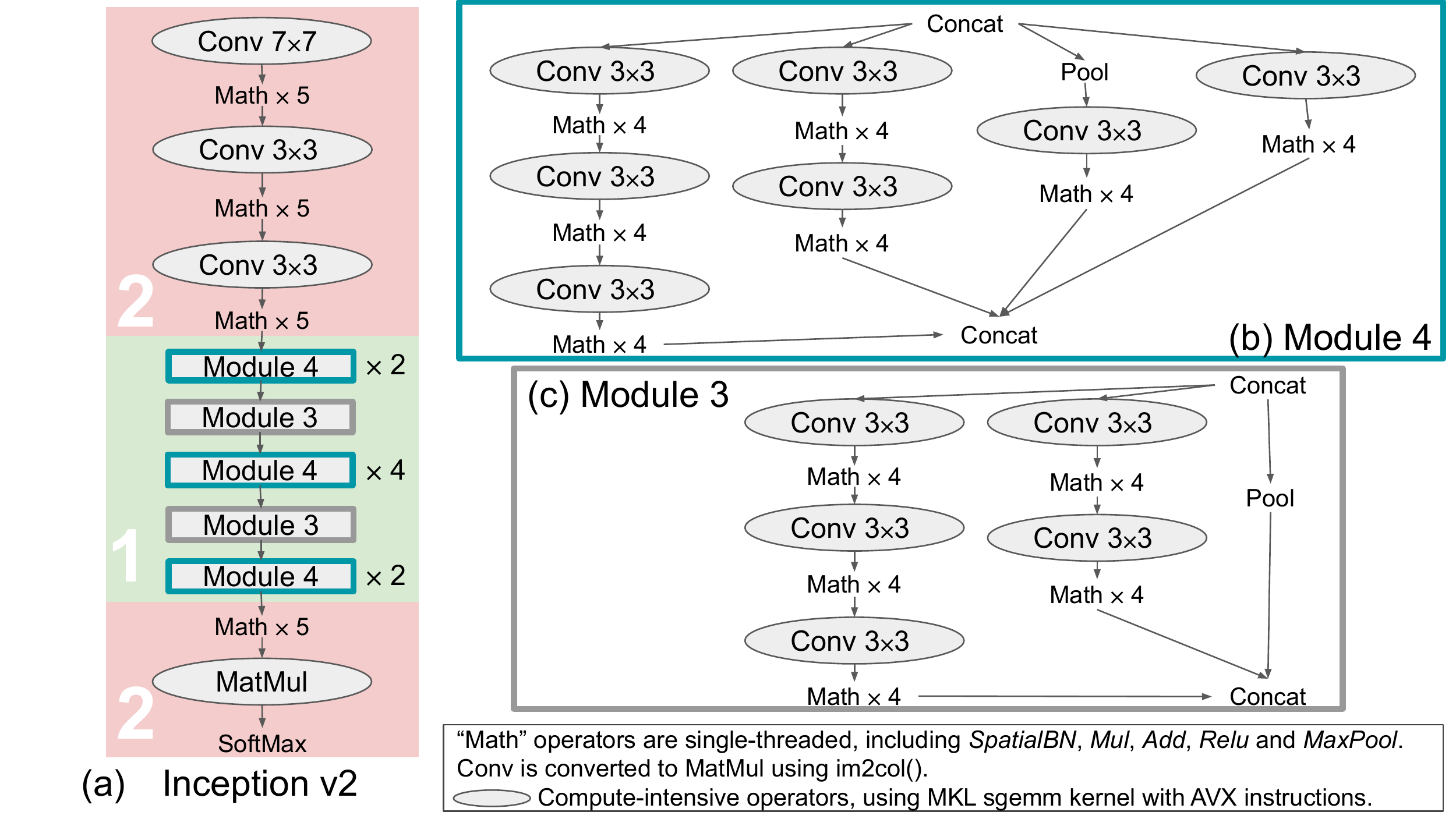}}

    \caption{(a) Inception v2 architecture contains modules with four (b) and three (c) independent branches.
    Area 1 exhibits inter- and intra-op parallelism and
    area 2 only intra-op.}

    \label{fig:sched:inceptionarch}
\end{figure}

\paragraph{Inception v2 Architecture}
To simplify the explanation of later results, we first summarize the Inception v2 architecture~\cite{szegedy2016rethinking} in \Fig{fig:sched:inceptionarch}.
\Fig{fig:sched:inceptionarch}a shows the top level architecture, color coded as areas 1 and 2.
Area 1 exhibits both intra- and inter-op parallelism, while area 2 only has intra-op parallelism.
Area 1 contains 
two inception modules.
Module 4 has four branches (\Fig{fig:sched:inceptionarch}b) and module 3 has three (\Fig{fig:sched:inceptionarch}c).
The convolution operators are converted to MatMul using \texttt{im2col()},
so intensive computation is mostly MatMul.

\begin{figure}[t]
    \centering
    \subfloat{\includegraphics[width=0.4\columnwidth]{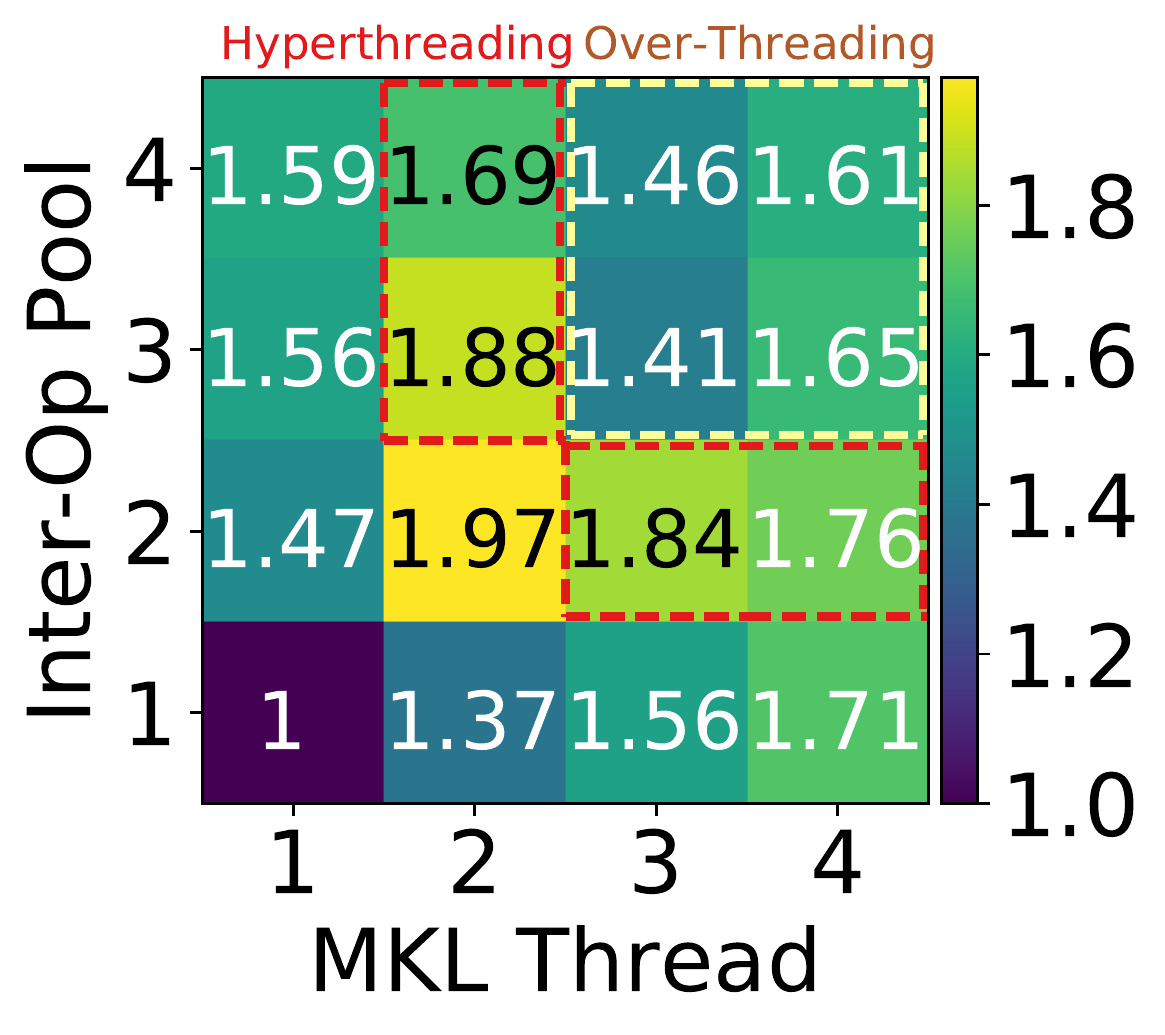}}

    \caption{Performance of Inception v2 with different numbers of inter-op pools and MKL threads per pool.
    Best configuration balances intra- and inter-op parallelism.
    }

    \label{fig:sched:threads}
\end{figure}

\paragraph{Performance Scaling with Pools and Threads}
\Fig{fig:sched:threads} shows the relative performance of Inception v2 with a batch size of 16, sweeping inter-op pools and MKL threads per pool.
The total number of threads on the system is the product of the two.
Hyperthreads are used when more than four threads are created.
Exceeding eight threads is labeled over-threading because there are more software threads than hardware threads. (Scaling is similar with batch sizes from 1 to 128.)

Hyperthreading does not improve performance significantly, such as  [4,1] vs [4,2], and [1,4] vs [2,4] ([Threads, Pools]).
The compute-intensive operators, Convs and MatMuls, are bottlenecked by the fused multiply-accumulate (FMA) units, which are shared between hyperthreads on the same core.

As expected, over-threading, i.e., using more software threads than hardware threads, hurts the performance, because threading overhead increases with more software threads, and computing resources are saturated.
As a result, simply setting all framework knobs to the maximum does not yield the best performance. 

\begin{figure}[t]
    \centering
    \subfloat{\includegraphics[width=0.5\columnwidth]{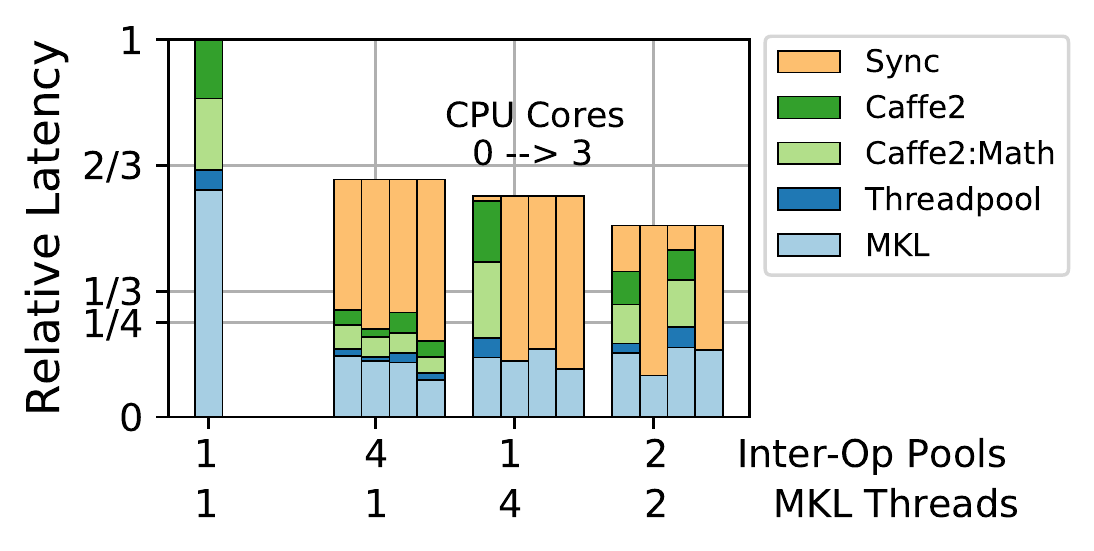}}
    \caption{Execution time breakdown of four cases.
    }
    \label{fig:sched:stackbar}
\end{figure}

\begin{figure}[t]
    \captionsetup[subfloat]{captionskip=-0.1em}
    \centering
    \subfloat[Execution traces of 4 inter-op pools, 1 MKL threads]{\includegraphics[width=0.6\columnwidth]{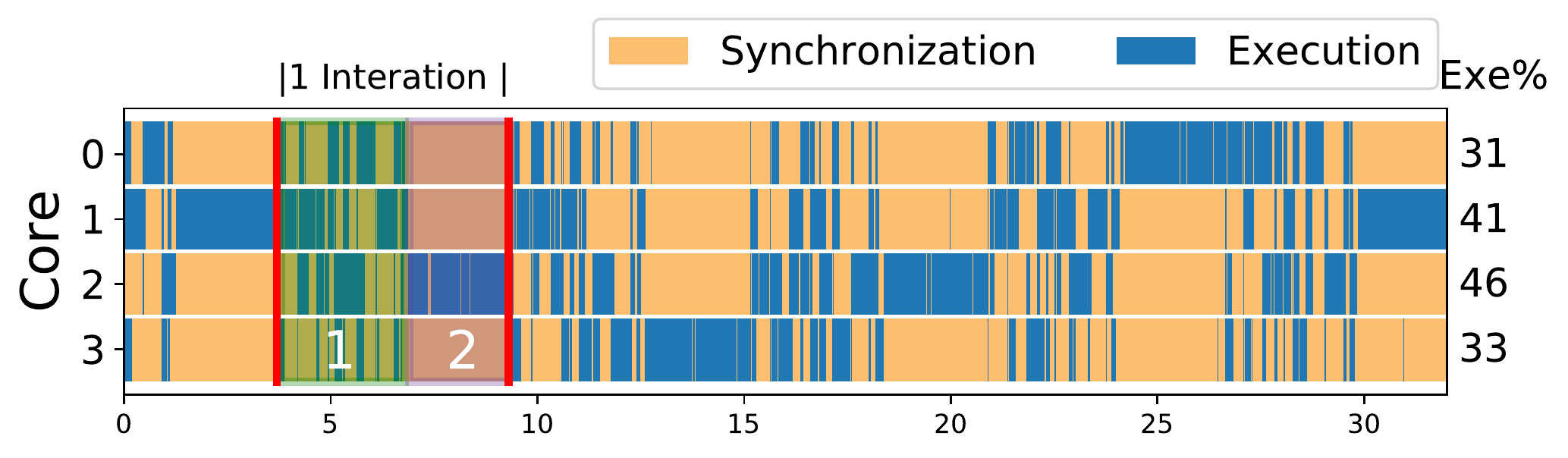}}
    \qquad
    \subfloat[Execution traces of 1 inter-Op pools, 4 MKL threads]{\includegraphics[width=0.6\columnwidth]{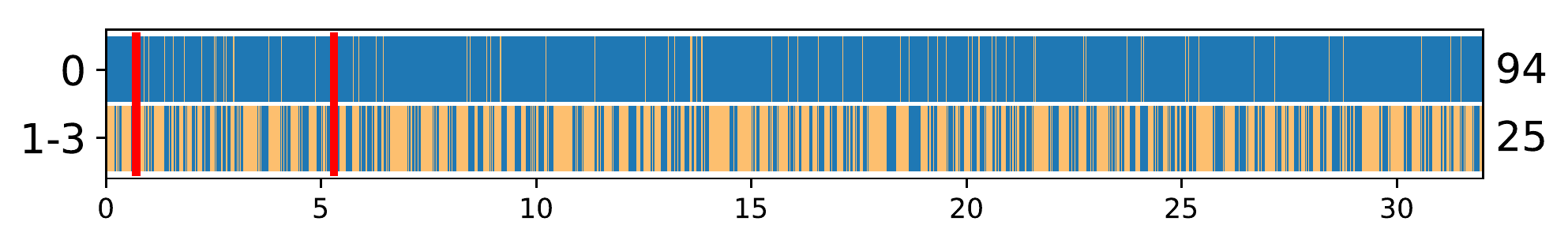}}
    \qquad
    \subfloat[Execution traces of 2 inter-op pools, 2 MKL threads]{\includegraphics[width=0.6\columnwidth]{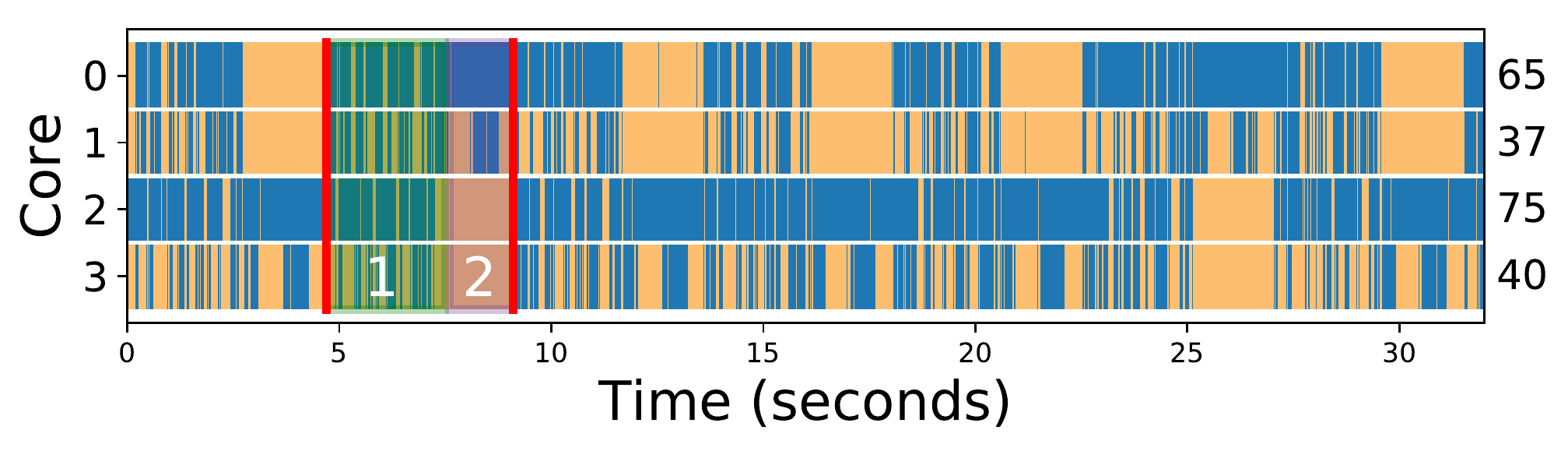}}
    \caption{Execution traces of three cases in \Fig{fig:sched:stackbar}.
    Color-coded areas 1 and 2 correspond to the operators in \Fig{fig:sched:inceptionarch}.
    }
    \label{fig:sched:traces}
\end{figure}

Performance is best with two pools and two threads per pool.
Using four total threads in other ways, such as four pools with one thread each, or one pool with four threads, cannot achieve such performance.
Our profiling methodology reveals and visualizes the underlying causes in Figures~\ref{fig:sched:stackbar} and~\ref{fig:sched:traces}.

\paragraph{Run-Time Breakdown and Execution Traces}
We select four cases, a baseline that uses only one thread, and three cases that each use four threads in total.
One software thread is bound to one CPU core.
\Fig{fig:sched:stackbar} shows the aggregate time breakdown, and \Fig{fig:sched:traces} shows the corresponding execution traces.
In \Fig{fig:sched:traces},
one iteration of Inception v2 inference is marked with red bars, and operators in execution are labeled with the corresponding color in \Fig{fig:sched:inceptionarch}, where area 1 exhibits intra- and inter-op parallelism, and area 2 has only intra-op parallelism.
The fraction of time each core spent executing (rather than synchronizing) is indicated to the right of each trace. It matches the breakdowns in \Fig{fig:sched:stackbar}.

The first case, using four pools of size one, incurs high synchronization overhead in \Fig{fig:sched:stackbar} and \Fig{fig:sched:traces}a, primarily because the operators in area 2, with only intra-op parallelism, are assigned only one core (thread), and other cores are waiting to synchronize.
The trace in area 1 is slightly better, with every core executing one branch of each inception module.

The second case, using one pool with four threads, does not perform well either, because the Caffe2-native operations (the parts labeled as Caffe2 and Caffe2:Math in \Fig{fig:sched:stackbar}) are single-threaded, and other cores are stalled by core 0, as shown by the long synchronization time in the rightmost three bars of \Fig{fig:sched:stackbar} and the traces of cores 1--3 in \Fig{fig:sched:traces}b.

The third case, using two pools, each with two threads, is a better balance.
It reduces synchronization time in both area~1 and area 2 compared to the first case.
Area 1 is improved because the number and size of convolutions in each inception branch is not even, as shown by \Fig{fig:sched:inceptionarch}b,
and allocating one core for each branch (the first case) makes the core running the smallest branch wait for a long time.
Area 2 is improved because its operators are sequential and compute-intensive, and having more cores improves performance.

\paragraph{Optimization Opportunity}
Operators in a complex model come in different sizes and have different dependencies.
Fixing each thread pool size usually incurs synchronization overhead because of work imbalance.
Thus there is an opportunity to implement a global thread pool, allowing the scheduler to determine dynamically how many threads to schedule for each operator.
For example, in the traces of \Fig{fig:sched:inceptionarch}, providing area 1 with two pools of two threads each and area 2 with one pool of four threads can lead to higher performance.

\section{Operator Design}
\label{sec:operator}

Operators are building blocks of DL frameworks, providing basic semantics
through high-level language (like Python) APIs to ease the development process for framework users.
As shown in \Fig{fig:motivation}, a workload built with a DL framework involves library kernels and framework native computation.
In this section, we show that efficient operator design can speed up framework native computation, and yields up to 4.2$\times$ performance improvement for real-world models.

\begin{figure}[t]
    \centering

    \subfloat{\includegraphics[width=0.4\columnwidth]{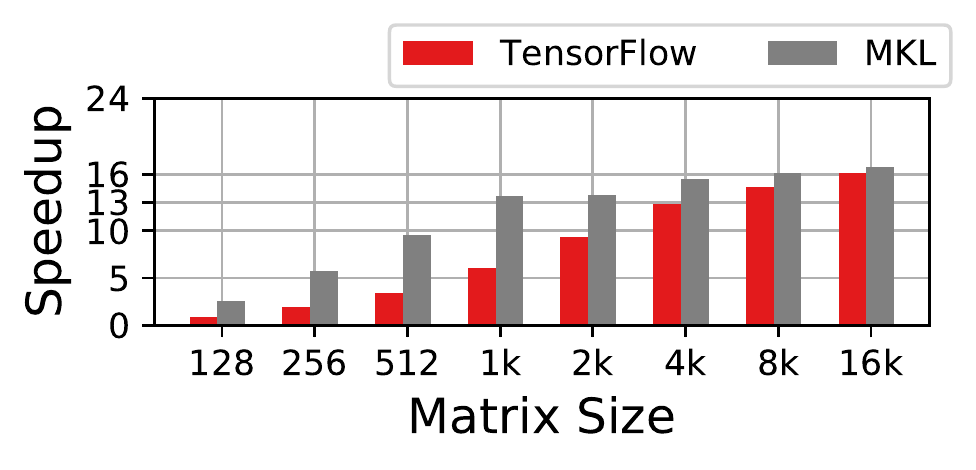}}

    \caption{Speedup from using 24 MKL threads instead of one.
    TensorFlow exhibits lower speedups than MKL.
    }

    \label{fig:op:threadscaling}
\end{figure}

\paragraph{Implementations of Framework Native Operators}
We first describe framework native operators, the operators that do not use library kernels.
Native operators handle control flow as well as tensor reshaping, broadcasting, and preprocessing.
Some, like those for control flow or input image preprocessing, are necessary. Others can fairly be described as a framework programmability tax.
The overhead stems from preparing inputs for library kernels, or computing how to parallelize a given workload in the main thread.
One example of the latter kind is Eigen::ParallelFor, used by TensorFlow.

\paragraph{Implementations of Compute-Intensive Operators}
A framework operator must sometimes do more than simply pass arguments to library kernels. Data preparation is often required, for example.
Taking matrix multiplication (MatMul) as an example, we list two implementations below.
In the context of deep learning, $x$ is an input matrix of size [batch size $\times$ number of
activations] and $w$ is a weight matrix of size [activations in current layer $\times$
activations in next layer].
We assume MatMul is the interface of a framework operator, and MATMUL is the corresponding library
kernel, e.g., in MKL.

\begin{verbatim}
Sec 5.1)  MatMul1(x, w):
            data_prep(x, w)
            return MATMUL(x, w)
\end{verbatim}

\begin{verbatim}
Sec 5.2)  MatMul2(x, w):
            // Reshape x, w into bx and bw
            for bx, bw in x, w:
                threadpool.run(MatMul1,
                bx, bw)
            return threadpool.join_results()
\end{verbatim}

MatMul1 conducts data preparation and passes the matrices to the library kernel.
MatMul2 uses an additional thread pool that we call the intra-operator thread pool, to distinguish it from the MKL thread pool.
The operator splits the matrices into smaller ones, and passes those small matrices to the intra-op thread pool.
The intra-op thread pool then executes multiple copies of MatMul1 in parallel.
This way data\_prep() of the whole matrix can be parallelized.
Before and after the library call, the data formatting and results gathering work is part of the programmability tax.
We will study the two implementations in the following subsections.

\subsection{MKL Threads}
\label{sec:op:mkl}

\begin{figure}[t]
    \centering
    \subfloat{\includegraphics[width=.5\columnwidth]{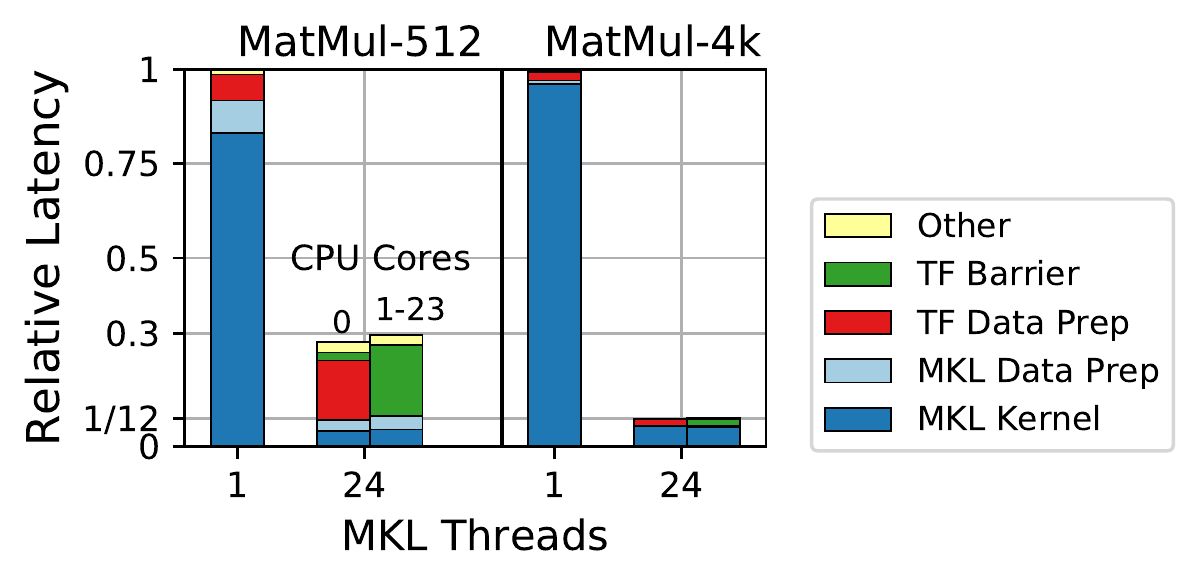}}
    \caption{Run-time breakdown for all CPU cores.
    Data preparation overhead causes the poor scalability in \Fig{fig:op:threadscaling}.
    }
    \label{fig:op:samples_mkl}
\end{figure}

By analyzing the overhead and scalability of the first operator implementation, MatMul1, we show that both the TensorFlow (TF) operator and the MKL kernel suffer from data preparation overhead, which prevents them from scaling linearly with the number of CPU cores.
The results here can also apply when convolution operators are converted to MatMuls using \texttt{im2col()}.
We use the \textit{large} platform in Table~\ref{table:cpu}.

\paragraph{Performance Scaling}
Both TF and MKL have scaling issues, and TF is slightly worse.
\Fig{fig:op:threadscaling} shows the speedup of using 24 MKL threads over using one, for both TF operators and MKL kernels.
The matrices are squared and represented by  one dimension. The total number of floating-point operations is the cube of that number.
\Fig{fig:op:threadscaling} shows that the speedup of TF is always lower than that of MKL, especially for small matrices.
TF speedup is comparable to MKL when matrices are larger than 4k.
The maximum speedup achievable is about 16$\times$, which is lower than the number of cores, 24.

\paragraph{Causes of the Poor Scalability}
Our profiling methodology reveals that
data preparation overhead causes suboptimal performance scaling.
We pick two variants of MatMul, MatMul-512 and MatMul-4k, that operate on medium- and large-size matrices, respectively.
MatMul-512 represents the fully-connected (FC) layers from YouTube~\cite{covington2016deep} and Facebook recommendation~\cite{naumov2019deep,gupta2019architectural} models, while MatMul-4k represents the FC layers in Transformer~\cite{vaswani2017attention}.
\Fig{fig:op:samples_mkl} shows the run-time breakdown of all CPU cores running the MatMuls, using 1 and 24 MKL threads.
With multiple threads, the thread tasked with lengthy TF data preparation is the main thread, labeled CPU Core 0.
The latency of each MatMul workload is normalized to that of using one MKL thread.

The TF parts of \Fig{fig:op:samples_mkl} show that TF's scaling issue is caused by framework overhead, due mainly to TF data preparation for MKL kernels.
Using one MKL thread, MatMul-512 spends over 10\% of its time in TF data preparation;
using 24 MKL threads, the overhead exceeds 72\%.
Overhead is much lower for MatMul-4k: less than 3\% in both cases.
Without TF overhead, speedup can clearly be much greater.
The MKL parts of \Fig{fig:op:samples_mkl} show that MKL data preparation causes the scaling issue for MKL kernels.
MKL kernel execution time is roughly 1/24 of the original run-time for MatMul-512.
Speedup drops with time spent in MKL data preparation.

\begin{figure}[t]
    \centering

    \subfloat{\includegraphics[width=0.7\columnwidth]{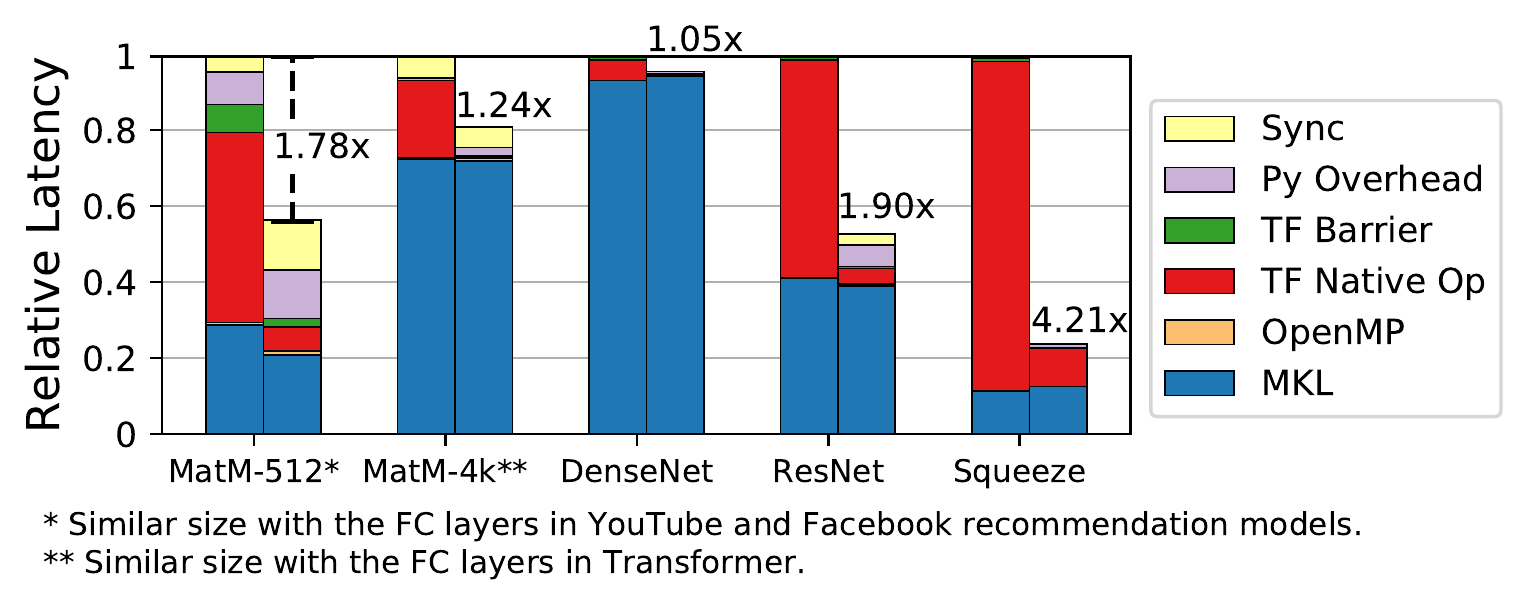}}

    \caption{Run-time breakdown of TensorFlow workloads with 1 (left bar) and 24 (right bar) intra-op threads.
    Both cases use 24 MKL threads.
    }

    \label{fig:op:intra_op}
\end{figure}

\paragraph{The Role of Framework Design}
The Amdahl's law bottleneck of DL frameworks is non-negligible.
The overhead a MatMul with size $n\times n \times n$ scales linearly with $n$ ($O$($n$)), while the number of floating-point operations scales cubically ($O$($n^3$)).
Thus the speedup of large MatMuls (e.g., 4k) is closer to ideal speedup.
Realistically, however, the most commonly-used FC layers are not always large enough.
Actually smaller ones are common in commercial workloads including YouTube's~\cite{covington2016deep} recommendation model (of size 256 to 1k) and Facebook's~\cite{naumov2019deep,gupta2019architectural} (of size 64 to 512).
Especially when hardware platforms are upgraded to higher floating-point computation capability, we need even larger matrices to amortize the overhead.
Thus it is key to focus optimization efforts on mitigating framework overhead.

\subsection{Intra-Operator Threads}
\label{sec:op:intraop}

After decades of optimizing the GEMM kernel, the performance bottleneck has shifted to the overhead of using such kernels, e.g., the data preparation overhead in \Fig{fig:op:samples_mkl}.
A natural approach to reducing the overhead in framework design is to parallelize  the framework native computation, with an intra-operator thread pool implemented at the framework level, as MatMul2 does.
We show that when library kernels are using FMA units, intra-op threads improve performance by utilizing other computational units on the same physical core, thereby benefiting from Intel's hyperthreading technology.
We use the \textit{large} platform from Table~\ref{table:cpu}.

\begin{figure}[t]
    \centering
    \subfloat{\includegraphics[width=0.6\columnwidth]{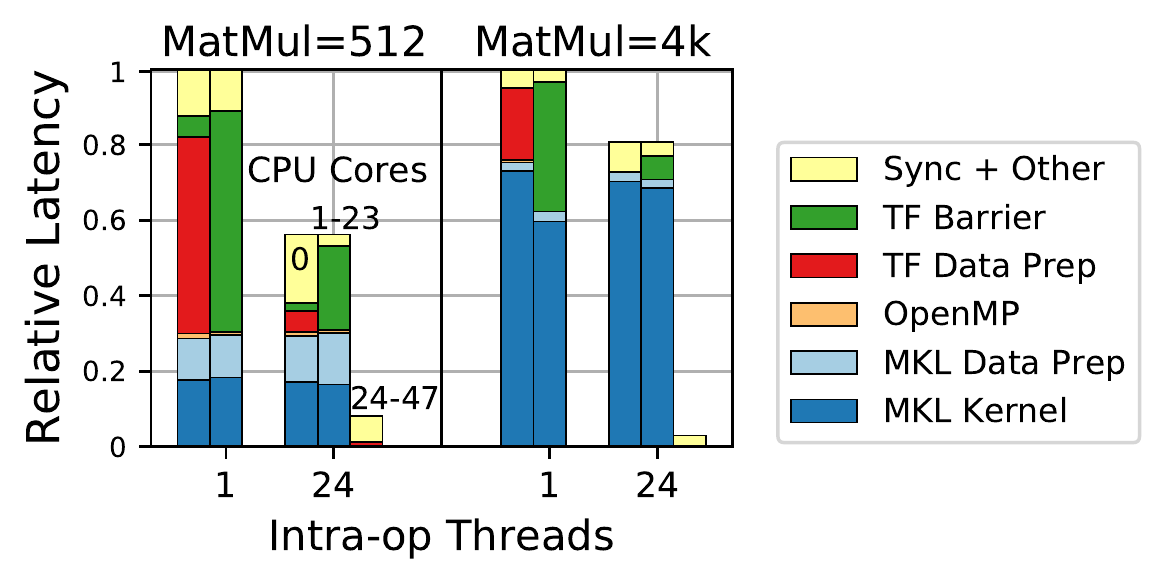}}
    \caption{Run-time breakdown of all CPU cores.
    Intra-op threads parallelize the overhead in cores 24-47.
    }
    \label{fig:op:perfsamples}
\end{figure}

\paragraph{Performance Improvement}
We first show how much performance improvement intra-op threads can yield and where it comes from.
\Fig{fig:op:intra_op} summarizes speedup and time breakdown when using 1 (left bar) and 24 (right bar) intra-op threads. Both cases use 24 MKL threads.
MatMul-512 and MatMul-4k are the same operators as in previous sections.
Using 24 intra-op threads reduces the execution time of TF native operators, while that of other parts stays similar.
The speedup ranges from 1.05$\times$ (DenseNet) to 4.21$\times$ (SqueezeNet).

Workloads bottlenecked by TF native operators benefit more from intra-op threads.
Such workloads, including MatMul-512 and SqueezeNet, have small to medium MatMul or convolution operations, because  TF native operators are likely to consume larger fractions of computation time.
For example, SqueezeNet has a small percentage of MKL computation since it is designed to have fewer parameters than AlexNet by using many small (1x1) convolution kernels~\cite{iandola2016squeezenet}.

\paragraph{Programmability Tax}
\Fig{fig:op:intra_op} also quantifies the framework programmability tax.
We estimate the tax using the non-MKL fractions, since they are not compute-intensive and can be largely optimized if written with high-performance language/code as MKL kernels.
After optimizing with intra-op threads, the programmability tax ranges from 1.3\% (DenseNet) to 63\% (MatMul-512).
SqueezeNet (47\%) is higher than ResNet-50 (26\%).
MatMul-4k (11\%) is slightly smaller.
This is the price we are paying for using frameworks.

\paragraph{Full-System Profiling}
Our profiling methodology visualizes the execution
of every core on the CPU platform to expose the reasons for performance improvement.
\Fig{fig:op:perfsamples} shows the time breakdown for all 48 hyperthreads on the \textit{large} platform.
We focus on the two MatMuls, since they are the simplest workloads.
Because cores 24-47 are not active when using one intra-op thread, the third bar is omitted for that case.
Core~0 of each case is the same as in \Fig{fig:op:intra_op}.

\Fig{fig:op:perfsamples} shows that with 24 intra-op threads, TF data preparation is distributed to cores 24 through 47. 
(The bottom of the third bar for MatMul-512 with 24 intra-op threads shows a tiny TF data preparation cost.)
That shortens TF data preparation time in core 0, so that the TF barrier time of cores~1 to 23 is shortened.
With only one intra-op thread, cores~1 to 23 spend about 60\% (MatMul-512) and 40\% (MatMul-4k) of time waiting in barrier, which is a big waste.

\paragraph{Hyperthreading}
Using intra-op threads takes advantage of Intel's hyperthreading technology
by colocating an intra-op thread and an MKL thread on the same physical core.
Since they need different hardware resources, they can execute in parallel without contention.
The critical path is the MKL thread. The intra-op thread adds no execution time to the overall workload.
For example, in \Fig{fig:op:perfsamples}, logical cores~0 and 24 are on the same physical core.
Core 0 executes mostly MKL floating-point operations with FMA units, which core 24 does not need.
Even without hyperthreading, the implementation of intra-op threads improves performance through parallelization.
In that case, the physical core's critical path combines the intra-op thread with the MKL thread.

\section{Library Choice}
\label{sec:library}

\begin{figure}[t]
    \centering
    \subfloat[Cycle Breakdown\label{fig:lib:top}]{\includegraphics[width=0.4\columnwidth]{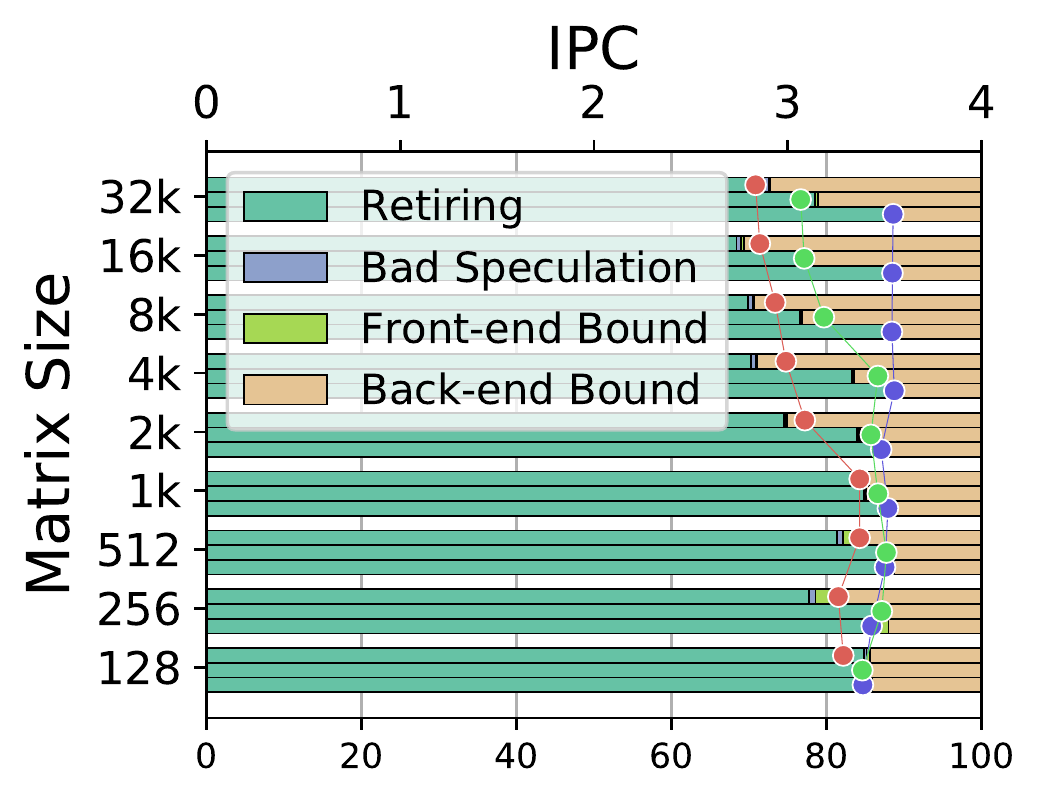}}
    \subfloat[LLC MPKI\label{fig:lib:llc}]{\includegraphics[width=0.15\columnwidth]{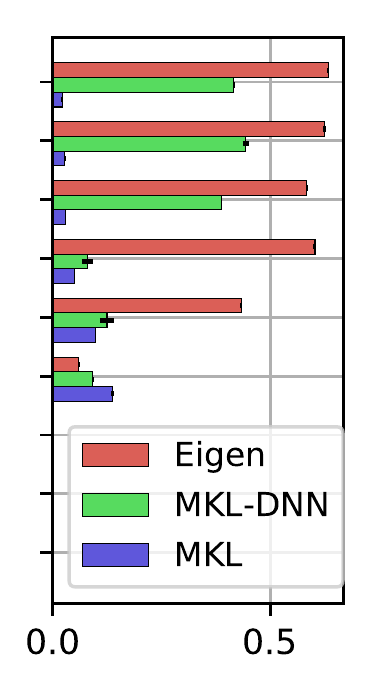}}
    \subfloat[BW (GB/s)\label{fig:lib:bw}]{\includegraphics[width=0.148\columnwidth]{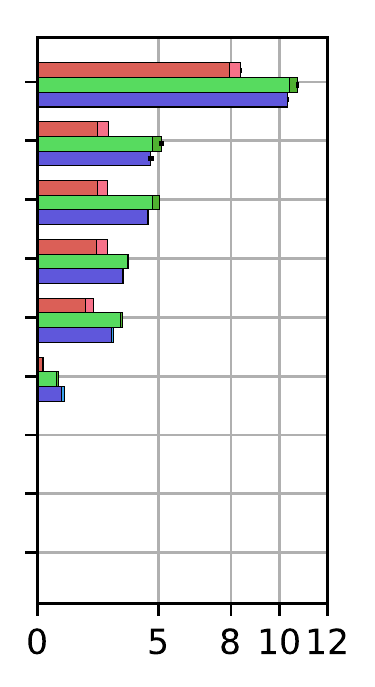}}

    \caption{(a) Cycle breakdown (bottom axis) and IPC (top axis)
    for three libraries multiplying matrices of various sizes. MKL has the highest retiring ratio and IPC,
    because it is the least back-end bound.
    (b) MKL has the lowest LLC miss rate. (c) The memory bandwidth consumption.
    MKL's prefetching is the most effective: almost all memory traffic is prefetching.
    }

    \label{fig:lib:toptop}
\end{figure}

Thanks to the mathematics and thread pool libraries, deep learning framework developers do not have to implement every basic functions from scratch.
In this section we study libraries for machine learning and thread pools.
We show that optimization can improve a GEMM kernel's performance by up to 25\%, owing to more efficient data prefetching.
We also find that robust thread pools such as Eigen
and Folly are better able to keep production-critical workloads running with little variation,
thus investing in sophisticated implementations is worthwhile for service providers.

\subsection{Machine Learning Library}

We compare MKL, MKL-DNN, and Eigen
with GEMM (general matrix multiplication) microbenchmarks on the \textit{small} platform (Table~\ref{table:cpu}) to expose architectural bottlenecks.

We conduct top-down analysis~\cite{yasin2014top} for single-threaded GEMM kernels with a variety of matrix sizes.
\Fig{fig:lib:top} shows the cycle breakdown (stacked bars) and IPC (dots).
The three bars shown for each matrix size are for Eigen, MKL-DNN, and MKL, from top to bottom.
For GEMM, MKL performs the best, followed by MKL-DNN.
With matrices larger than 4k, about 25\% of cycles are back-end bound for Eigen and MKL-DNN.

The back-end bottleneck is caused by last-level-cache (LLC) misses, shown as LLC misses per thousand instructions (MPKI) in \Fig{fig:lib:llc}.
Eigen and MKL-DNN have much higher LLC MPKI than MKL.
The LLC miss rate difference is caused by the aggressiveness and effectiveness of data prefetching, shown by
the memory traffic in \Fig{fig:lib:bw}, where the right ends of the bars show memory traffic incurred by LLC misses.
MKL's memory traffic is close to that of MKL-DNN, and its much lower LLC miss rate shows that MKL's software prefetching is more \textit{effective}.

We compare the GEMM kernels to demonstrate how and why kernel performance can vary.
MKL-DNN is likely outperform other libraries for other kernels, because
it is a library targeting deep learning with DL-specific optimizations.
It reportedly outperforms other frameworks~\cite{mkldnnopt}.

\subsection{Thread Pool Library}

We compare three thread pools, one simple implementation using std::thread,
and the thread pools in Eigen~\cite{eigenthreadpool} and Folly~\cite{folly}.
Our microbenchmark creates a thread pool
and starts $10k$ tasks that increment the value of a globally shared variable. This microbenchmark examines the case where the computation per thread is minimal and the synchronization between threads is the maximum. It is a common way to stress-test thread pools such as in Pytorch\footnote{\url{https://github.com/pytorch/pytorch/blob/master/binaries/at_launch_benchmark.cc}} as an isolation study.
We benchmark in two scenarios: (1) setting the number of threads to be the same as the number of
physical cores, and (2) using many more threads than the number of cores. 
We use the \textit{small} platform from Table~\ref{table:cpu}.

\begin{figure}[t]
    \centering

    \subfloat{\includegraphics[width=0.5\columnwidth]{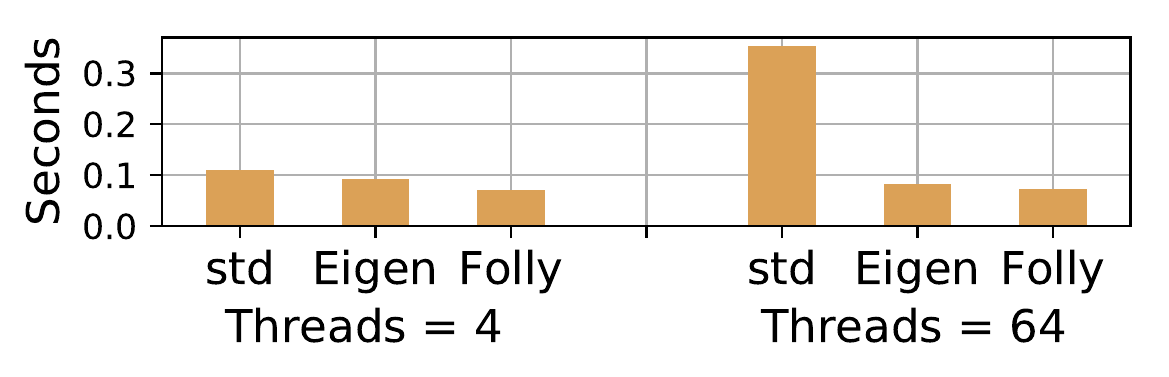}}
  
    \caption{Thread pool overhead, measured as time to run 10k micro tasks. Folly outperforms std::thread and Eigen even in an extreme case with 64 threads on a 4-core CPU.}

    \label{fig:lib:threadpool}
\end{figure}

\Fig{fig:lib:threadpool} compares the overall latency of running 10k micro tasks.
In both cases, Folly outperforms Eigen and Eigen outperforms std::thread.
When thread pool size is 64, greatly exceeding the number of CPU cores, Folly and Eigen perform consistently well as with four threads, and
oversubscribing the system does not drastically increase the overhead.
But the overhead of std::thread
increases by over 3$\times$, with
every CPU core spending about 60\% of its time in synchronization.

\section{Beyond One Socket}

In previous sections we have explored the framework designs on one-socket CPUs.
In this section, we study how those design features can be applied to scale out the workloads beyond one socket.
Unsurprisingly, the bottleneck of a two-socket system is the UPI bandwidth between sockets.
We study scaling one operator to two sockets and scheduling multiple operators to different sockets, as scaling-out versions of intra- and inter-op parallelism studies.
The experiments are conducted on the \textit{large} and \textit{large.2} platforms from Table~\ref{table:cpu}.

\begin{figure}[t]
    \centering
    \subfloat{\includegraphics[width=0.6\columnwidth]{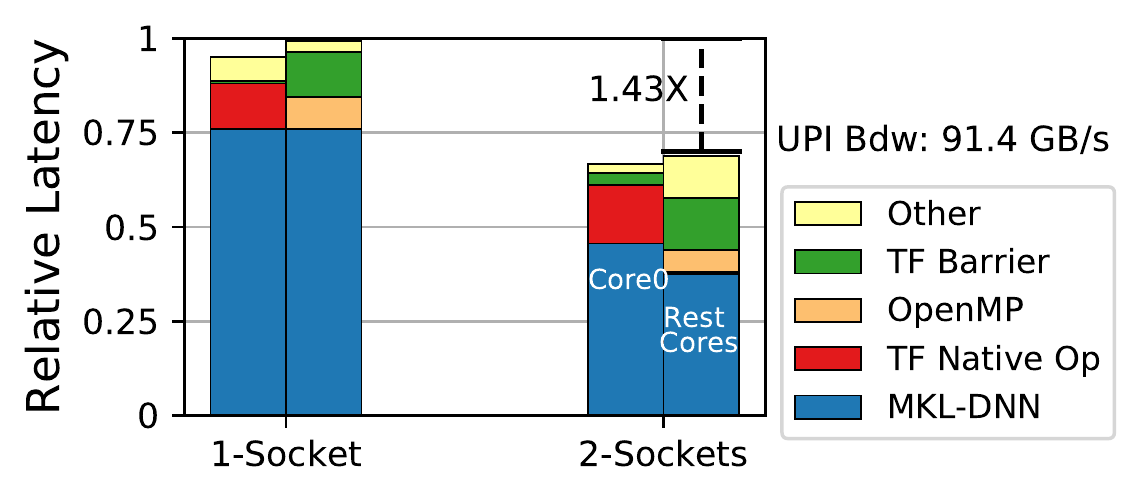}}

    \caption{A two-socket platform speeds up ResNet-50 by 1.43$\times$.
    The bottleneck is the UPI bandwidth, increasing TF data preparation time as part of the TF native operator time.}

    \label{fig:2sockets:resnetstackbars}
\end{figure}

\subsection{Data Parallelism}

We study data parallelism by setting the numbers of intra-op and MKL threads to the total number of physical cores and the number of inter-op threads to one.

\paragraph{ResNet Performance}
\Fig{fig:2sockets:resnetstackbars} shows the execution time breakdown of ResNet-50 running on one- and two-socket platforms. 
The latter speeds up ResNet by 1.43$\times$, less than the two-fold hardware increase.
The bottleneck is that UPI traffic peaks at 91.4GB/s, compared to
the theoretical maximum of 120GB/s.
UPI saturation increases the latency of TF native operators on a two-socket platform, which now includes both data preparation and transfer time between sockets.

\begin{figure}[t]
    \centering
    \captionsetup[subfloat]{captionskip=-0.1em}
    \subfloat[Two-Socket Speedup]{\includegraphics[width=0.3\columnwidth]{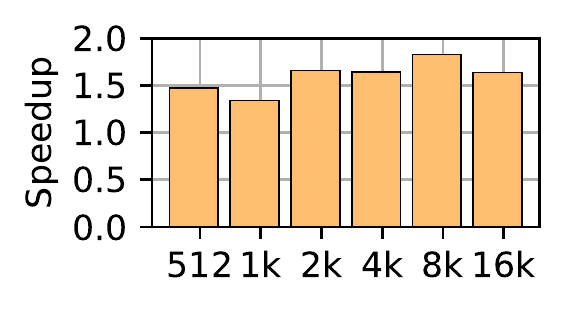}}
    \subfloat[UPI Traffic]{\includegraphics[width=0.3\columnwidth]{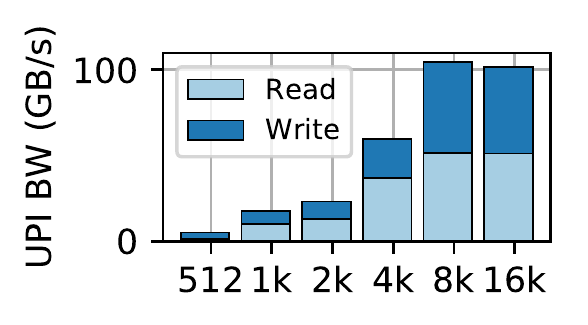}}

    \caption{(a) Speedup of a two-socket platform over one socket.
    (b) Measured peak UPI bandwidth consumption on the two-socket platform is close to 100GB/s.}

    \label{fig:2sockets:fc}
\end{figure}

\paragraph{MatMul Performance}
To test the limit of the \textit{large.2} platform,
we conduct microbenchmarking using TensorFlow MatMul operators.
Similarly, the UPI bandwidth is the bottleneck for large MatMuls.
\Fig{fig:2sockets:fc} shows two-socket speedup and corresponding UPI bandwidth consumption.
The speedup and UPI throughput increase with larger MatMul sizes, and peak for MatMul-8k.
For MatMul-16k, speedup decreases and bandwidth saturates, indicating empirically the maximum UPI bandwidth is around 100GB/s for such workloads.

\begin{figure}[t]
    \centering
    \subfloat{\includegraphics[width=0.7\columnwidth]{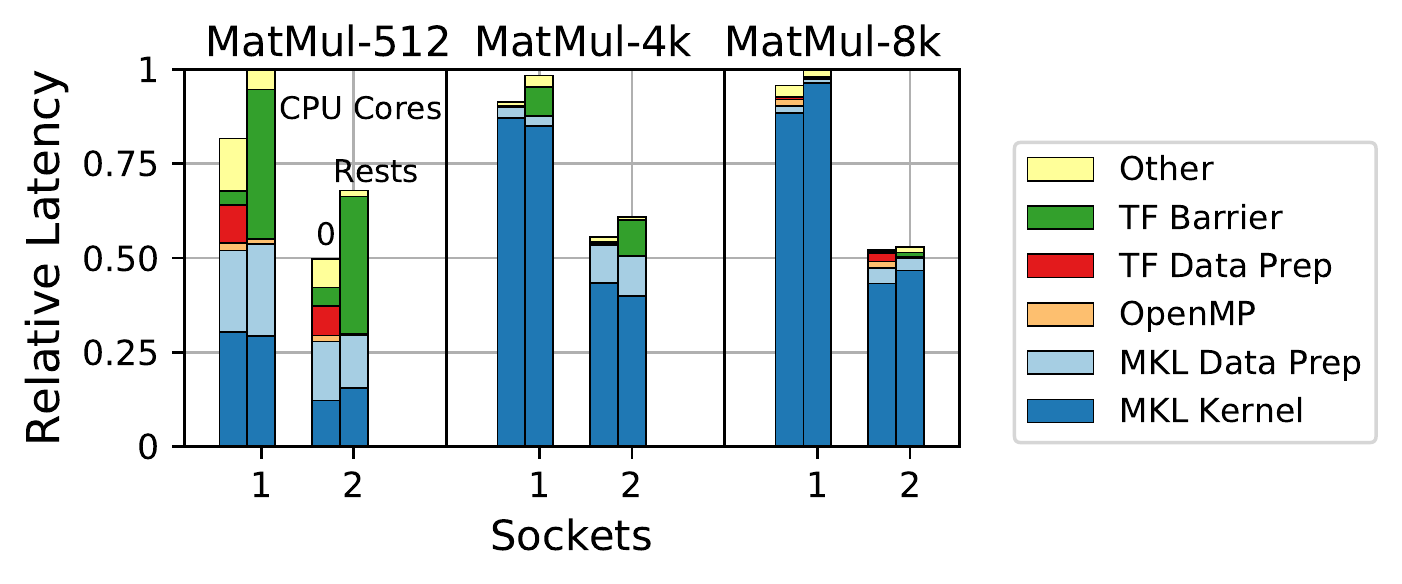}}

    \caption{Run-time breakdown of all CPU cores.
    }

    \label{fig:2sockets:fcstackbars}
\end{figure}

The speedup of a workload is determined by its intrinsic parallelism and UPI bandwidth saturation.
\Fig{fig:2sockets:fcstackbars} shows the time breakdown of MatMuls running on one- and two-socket platforms.
For medium MatMul sizes like 512, the poor scalability is caused by the limited parallelism of the workload, which cannot hide data preparation overhead.
For larger MatMuls, 4k and 8k, the data preparation time of both TF and MKL increases on the two-socket platform because of UPI saturation.
MatMul-8k has the best balance of intrinsic parallelism and UPI throughput.
It leads to the highest speedup (1.8$\times$, i.e., 44\% less execution time than with just one socket), which is close to perfect scaling.

\subsection{Model Parallelism}

We study model parallelism of a two-socket platform by using two inter-op pools, one per CPU socket.
Model parallelism improves performance significantly when the parallel operators are on critical paths and have similar sizes,
as with multiple embedding operators in neural collaborative filtering (NCF).
Performance and model parallelism mechanisms will be discussed in the next section.

Model parallelism does not always improve performance.
One example is the inter-op parallelism from training workloads, as in \Sec{sec:scheduler}.
Assigning gradient and weight sum operators one socket each causes workload imbalance between two sockets when batch size is large.
Two-socket platforms are not beneficial when the intra-op thread pool is not implemented at the framework level, since the workload bottleneck is single-threaded operators.

\section{Framework Design Tuning}
\label{sec:tuning}

At the outset, we identified five design features:  scheduling mechanism, operator implementation, math library, thread pool library, and the parallelism mechanism for platforms larger than one socket.
Our analysis has shown that 
the most effective setting for users to determine is the proper number of inter-op pools based on the model architecture.
Intra-op parallelism configurations follow from that setting.
Based on the analysis in previous sections, we therefore propose a set of simple yet effective guidelines.
The guidelines are architecture-irrelevant, because they are summarized based on software design analysis; thus they can be applied to CPUs with different architectures including Intel, AMD and OpenPOWER.

\paragraph{Definitions}
To present the guidelines, we first define a few terminologies.
\Sec{sec:sched:performance} and \Fig{fig:sched:speedup} mentioned the maximum width of a model graph.
The \textit{\textbf{average width}} of a model graph quantifies its inter-op parallelism, and we use it 
to determine the number of inter-op pools for a model.
The average width of a model is the floor of the ratio of the total number of (heavy) operators divided by the maximum number of layers.
A \textit{\textbf{heavy operator}} is a compute-intensive or embedding operator that usually takes significantly longer execution time than other operators.
Examples are the Conv operators in \Fig{fig:sched:inceptionarch}, as opposed to lightweight math operators, which are not considered.
The average model width of \Fig{fig:sched:inceptionarch}b is $\lfloor\frac{7}{3}\rfloor = 2$.

\paragraph{Guidelines}
The number of inter-op pools ($p$) is chosen to be the average model width. 
After $p$ is chosen, we choose the numbers of MKL and intra-op threads such that the entire system is split into $p$ partitions without redundant threads (\Sec{sec:scheduler:inception} and \Fig{fig:sched:threads}).
Therefore, the number of MKL threads and the number of intra-op threads for each thread pool should be equal to the total number of physical cores on the system divided by $p$.
That way one MKL thread and one intra-op thread can share the same physical core.
MKL threads can use the FMA units and the intra-op threads can use other units via hyperthreading (\Sec{sec:op:intraop} and \Fig{fig:op:perfsamples}).

\begin{figure}[t]
    \centering
    \subfloat{\includegraphics[width=0.8\columnwidth]{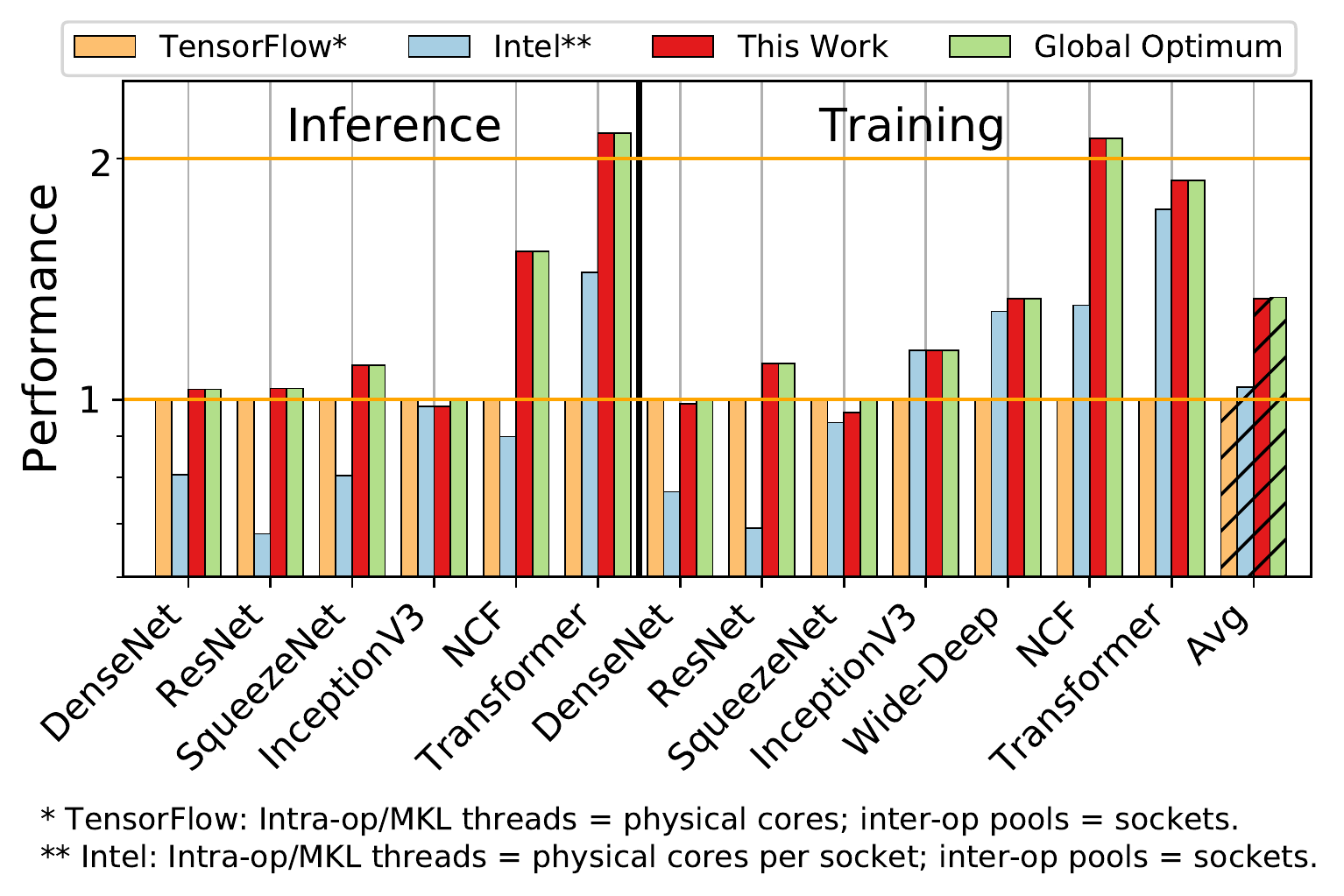}}
    \caption{Performance using the recommended TensorFlow settings~\cite{tensorflowperformance} (baseline),  Intel blog~\cite{inteltips}, our work, and global optimum obtained by exhaustive search.
    Our work outperforms Intel and Tensorflow suggestions and nearly closes the gap between the state of the art and the global optimum.}
    \label{fig:speedup}
\end{figure}

\begin{table}[t]
\small
\begin{center}
\begin{tabular}{|c|c|c|c|c|c|c|c|c|c|}
\hline
 Dense & Squeeze & ResNet & IncepV3 & W/D & NCF & Trans \\ 
\hline
1 & 1 & 1 & 2 & 3 & 4 & 4\\
\hline
\end{tabular}
\end{center}
\caption{Average model width, i.e., the number of pools selected for \Fig{fig:speedup} based on our guidelines.
Intra-op and MKL threads $=$ total physical cores divided by those numbers.}
\label{table:tuning}
\end{table}

\paragraph{Evaluation Setup}
We integrate our guidelines with TensorFlow v1.13. (TensorFlow refers to inter-op pools as inter-op parallelism threads.)
We will open source the TensorFlow plugin to the public, to let users apply our guidelines automatically.
The size of the design space encompassing the numbers of MKL, intra- and inter-op threads is the cube of the number of logical cores. For the \textit{large.2} system, that means $96^3 = 884,736$ design points. Our guidelines suggest picking only \textit{one} of those $884,736$ possibilities.

We evaluate our guidelines by applying the rules to a fresh set of workloads and by performing the evaluation on a different platform from that used to develop the guidelines.
Our analysis uses microbenchmarks and vision models that run with images;
for evaluation, we add Inception v3~\cite{szegedy2016rethinking}, the wide-deep recommendation model~\cite{widedeep}, the neural collaborative filtering model (NCF)~\cite{he2017neural}, and Transformer~\cite{vaswani2017attention}, covering recommendation and translation workloads.
The bulk of our analysis uses platforms \textit{small} and \textit{large} from Table~\ref{table:cpu};
here we evaluate the guidelines on the \textit{large.2} platform, the largest AWS bare metal instance.

\paragraph{Speedup}
\Fig{fig:speedup} summarizes the speedup of this work over TensorFlow~\cite{tensorflowperformance} (baseline) and Intel~\cite{inteltips} recommended settings.
It also compares our settings' performance to the global optimum obtained by exhaustively sweeping the design space. 
TensorFlow suggests setting the number of MKL and intra-op threads to the physical core count, and inter-op pools to the socket count.
Intel suggests setting MKL and intra-op threads to the number of physical cores per socket, and inter-op pools to the socket count.
Our analysis shows that TensorFlow suggests more threads than needed, and Intel's setting is suitable for models with an average width of two.

Overall, our performance guidelines perform consistently better than the settings recommended by Intel and TensorFlow. 
Our method bridges the performance gap between those state-of-the-art settings and the global optimum for all evaluated workloads except Inception inference and SqueezeNet training.
In those two cases, our guidelines achieve 95\% of the performance offered by the global optimal setting.
On average, this work achieves
the same performance as the global optimum, and
1.34$\times$ and 1.29$\times$ better performance than TensorFlow's and Intel's suggestions, respectively.

We summarize the average model width in Table~\ref{table:tuning}. It is the same as the number of inter-op pools in use.
The models shown have average width between one and four, which is diverse.
The numbers of MKL and intra-op threads for each model is the total number of physical cores (48) divided by the model width.
For example, the setting for the W/D (wide and deep) model is 3 inter-op pools, 16 MKL threads, and 16 intra-op threads,
which is also the globally optimal setting.

The performance guides from Intel and TensorFlow are general, aiming to make it easy for most users to get reasonable performance, and they perform reasonably well.
For vision models, TensorFlow's settings perform as well as the global optima and our guidelines, while Intel's does not perform well for the vision models except for Inception, because Intel's setting favors models with inter-op parallelism that other vision models do not have.
Intel's settings perform better than TensorFlow's for recommendation and translation models. The latter have several parallel embedding operators, thus their average width is no less than two.
The default TensorFlow setting performs much worse than both the Intel and TensorFlow recommendations. TensorFlow naively sets all parameters---MKL threads, intra-op threads, and inter-op pools---to the number of logical cores. As pointed out in our earlier analysis, this is sub-optimal.
Thus TensorFlow users who run only one model and one session at a time
should at least set the number of inter-op pools to one instead of using the default setting.

\section{Conclusion}
We presented a detailed evaluation and analysis of key design features and the role of parallelism in a machine learning framework, focusing on scheduling, operator implementation, and library back ends.
To maximize parallelism, we proposed simple guidelines for tuning framework parameters, distilled from detailed domain-specific design feature knowledge and analysis. 
We demonstrated the usability and the additional performance improvement of this approach by integrating and evaluating our methodology with TensorFlow.
On average, our method outperformed the suggested settings from Intel and TensorFlow performance guides by 1.29$\times$ and 1.34$\times$, respectively, across a set of real-world DL models.

\bibliographystyle{ACM-Reference-Format}
\bibliography{main}

\end{document}